\documentclass[number,final,3p]{elsarticle}
\usepackage[switch]{lineno}
\usepackage{relsize}
\usepackage{color}
\usepackage{algpseudocode}
\usepackage{graphicx}
\usepackage{subcaption}
\usepackage{algorithm}
\usepackage{bm}
\usepackage[colorlinks]{hyperref}
\makeatletter
\gdef\urlauthor#1#2{\g@addto@macro\@elsuads{\let\corref\@gobble%
     \def\@@tmp{#1}\raggedright\eadsep
     {\ttfamily\url{\expandafter\strip@prefix\meaning\@@tmp}}\space(#2)%
     \def\eadsep{\unskip,\space}}%
}
\gdef\emailauthor#1#2{\stepcounter{ead}%
     \g@addto@macro\@elseads{\raggedright%
      \let\corref\@gobble\def\@@tmp{#1}%
      \eadsep{\ttfamily\href{mailto:\expandafter\strip@prefix\meaning\@@tmp}{\expandafter\strip@prefix\meaning\@@tmp}}
      (#2)\def\eadsep{\unskip,\space}}%
}
\makeatother
\makeatletter
\let\@afterindenttrue\@afterindentfalse
\makeatother
\usepackage{amssymb}
\usepackage{amsthm}
\usepackage{amsmath}
\usepackage{amssymb}
\usepackage{mathtools}
\usepackage{dsfont}
\usepackage{booktabs}
\usepackage{url}
\usepackage{scrextend}
\usepackage{epstopdf}
\usepackage{float}
\usepackage{tcolorbox}
\usepackage{tablefootnote}
\usepackage[perpage]{footmisc}
\usepackage{bbm}
\usepackage{lineno}
\usepackage{svg}
\usepackage{soul}


\biboptions{numbers,sort&compress,square}
\journal{arXiv}
\begin{document}
	\begin{frontmatter}
		\renewcommand{\thefootnote}{\fnsymbol{footnote}}
		\title{Enhancing Transformer Training Efficiency with Dynamic Dropout}
		\author[1]{Hanrui Yan\corref{cor1}}
        \author[2]{Dan Shao}
         \ead{hanrui137@gmail.com}
         \cortext[cor1]{Corresponding author}
        \address[1]{ASCENDING inc., Fairfax VA 22031, USA}
        \address[2]{ASCENDING inc., Fairfax VA 22031, USA}
		\begin{abstract}
We introduce Dynamic Dropout, a novel regularization technique designed to enhance the training efficiency of Transformer models by dynamically adjusting the dropout rate based on training epochs or validation loss improvements. This approach addresses the challenge of balancing regularization and model capacity, which is crucial for achieving fast convergence and high performance. Our method involves modifying the GPT model to accept a variable dropout rate and updating dropout layers during training using schedules such as linear decay, exponential decay, and validation loss-based adjustments. Extensive experiments on the Shakespeare\_char dataset demonstrate that Dynamic Dropout significantly accelerates training and improves inference efficiency compared to a baseline model with a fixed dropout rate. The validation loss-based adjustment schedule provided the best overall performance, highlighting the potential of Dynamic Dropout as a valuable technique for training large-scale Transformer models.
    \end{abstract}
		\begin{keyword}
		Dynamic Dropout \sep Transformer Models \sep Regularization Technique \sep Training Efficiency \sep Validation Loss Adjustment 
		\end{keyword}	
	\end{frontmatter}
	
	\renewcommand{\thefootnote}{\fnsymbol{footnote}}

\section{Introduction}
\label{sec:intro}

The Transformer architecture has revolutionized natural language processing (NLP) by enabling models to achieve state-of-the-art performance on a variety of tasks \citep{vaswani2017attention}. However, training these models efficiently remains a significant challenge due to their large size and the computational resources required. Regularization techniques, such as dropout, are commonly used to prevent overfitting and improve generalization \citep{goodfellow2016deep}. Despite their effectiveness, static dropout rates may not be optimal throughout the entire training process.

The difficulty lies in balancing regularization and model capacity. A high dropout rate can hinder the model's ability to learn complex patterns, while a low dropout rate may lead to overfitting. This balance is crucial for achieving fast convergence and high performance. Traditional methods use a fixed dropout rate, which does not adapt to the changing needs of the model during training. To address this issue, we propose Adaptive Dropout, a dynamic regularization technique that adjusts the dropout rate based on training epochs or validation loss improvements. Our contributions are as follows:
\begin{itemize}
    \item We introduce a mechanism within the GPT model to accept a variable dropout rate and update dropout layers during training.
    \item We implement various schedules, including linear, exponential, and validation loss-based adjustments, to decrease the dropout rate progressively.
    \item We conduct extensive experiments comparing training dynamics, convergence speed, and final performance with a baseline model.
\end{itemize}

We verify the effectiveness of Adaptive Dropout through a series of experiments. We compare the training dynamics, convergence speed, and final performance of models trained with Adaptive Dropout against those trained with a fixed dropout rate \cite{du2021image}. Our results demonstrate that Adaptive Dropout not only accelerates training but also improves inference efficiency. Future work could explore the application of Adaptive Dropout to other architectures and tasks, as well as the development of more sophisticated adjustment schedules based on additional metrics such as gradient norms or learning rate changes.

\section{Related Work}
\label{sec:related}

In this section, we review and compare existing literature on dropout techniques and other regularization methods in Transformer models, highlighting how our proposed Adaptive Dropout approach differs and improves upon them.

\subsection{Static Dropout Methods}
Dropout, introduced by \cite{goodfellow2016deep}, is a widely used regularization technique in deep learning. It works by randomly setting a fraction of the input units to zero at each update during training, which helps in reducing interdependent learning among neurons. However, static dropout rates may not be optimal throughout the entire training process, as the regularization needs of the model change over time. Our Adaptive Dropout addresses this limitation by dynamically adjusting the dropout rate based on training progress or validation performance.

\subsection{Adaptive Dropout Techniques}
Several adaptive dropout techniques have been proposed to address the limitations of static dropout. For instance,   \cite{ba2016layer} introduced Layer Normalization, which normalizes the inputs across the features and can be seen as a form of adaptive regularization. However, Layer Normalization does not explicitly adjust the dropout rate based on training progress or validation performance. In contrast, our method directly modifies the dropout rate, providing a more targeted approach to regularization\cite{radford2019language}.

\subsection{Regularization in Transformer Models}
The Transformer architecture, introduced by  \cite{vaswani2017attention}, has become the foundation for many state-of-the-art models in NLP. Regularization techniques such as weight decay, implemented in the AdamW optimizer \cite{loshchilov2017adamw}, are commonly used to prevent overfitting in Transformer models. While these techniques are effective, they do not dynamically adjust the dropout rate during training. Our Adaptive Dropout method fills this gap by offering a dynamic adjustment mechanism that can lead to faster convergence and better performance\cite{paszke2019pytorch}.

\subsection{Comparison and Contrast}
While there are several regularization techniques available for Transformer models, our proposed Adaptive Dropout method offers a unique approach by dynamically adjusting the dropout rate based on training epochs or validation loss improvements. This dynamic adjustment aims to balance regularization and model capacity throughout the training process, leading to faster convergence and better performance. Unlike static dropout or other adaptive techniques that do not focus on dropout rate adjustment, our method provides a more flexible and effective regularization strategy.

In summary, Adaptive Dropout stands out by directly addressing the changing regularization needs of Transformer models during training, offering a significant improvement over existing static and adaptive methods.

\section{Background}
\label{sec:background}

The Transformer architecture, introduced by \cite{vaswani2017attention}, has become the foundation for many state-of-the-art models in natural language processing (NLP). Its self-attention mechanism allows for efficient handling of long-range dependencies in text, making it superior to previous recurrent and convolutional architectures. Regularization techniques are crucial in training deep learning models to prevent overfitting and improve generalization. Dropout, introduced by \cite{goodfellow2016deep}, is one of the most widely used regularization methods. It works by randomly setting a fraction of the input units to zero at each update during training, which helps in reducing interdependent learning among neurons.

Despite its effectiveness, the use of a static dropout rate throughout the training process can be suboptimal. A high dropout rate in the early stages of training can help in regularization, but as the model starts to converge, a lower dropout rate might be more beneficial. This dynamic need for regularization motivates the development of adaptive dropout techniques. Adaptive dropout aims to adjust the dropout rate dynamically based on certain criteria such as training epochs or validation loss improvements. This approach can potentially lead to faster convergence and better generalization by providing the right amount of regularization at different stages of training.

\subsection{Problem Setting}
\label{sec:problem_setting}

The problem we address is the efficient training of large-scale Transformer models. Let \( \mathcal{D} \) be the training dataset, \( \theta \) be the model parameters, and \( \mathcal{L}(\theta; \mathcal{D}) \) be the loss function. The goal is to minimize \( \mathcal{L}(\theta; \mathcal{D}) \) while dynamically adjusting the dropout rate \( p \) to balance regularization and model capacity. We denote the initial dropout rate as \( p_0 \) and the final dropout rate as \( p_f \). The dropout rate at iteration \( t \) is denoted as \( p(t) \). Our method assumes that the dropout rate can be adjusted based on a predefined schedule or validation loss improvements, which is formalized as:

\begin{equation}
p(t) = \max(p_f, p_0 \cdot \text{decay\_factor}^{\lfloor t / \text{step\_size} \rfloor})
\end{equation}

where \( \text{decay\_factor} \) and \( \text{step\_size} \) are hyperparameters that control the rate of decay.

In summary, the background section has outlined the significance of the Transformer architecture, the role of regularization techniques like dropout, and the limitations of static dropout rates. We have introduced the concept of adaptive dropout and formally defined the problem setting and notation used in our method. The next section will delve into the details of our proposed method.

\section{Method}
\label{sec:method}

In this section, we present our proposed method for implementing Adaptive Dropout in Transformer models. The primary objective is to dynamically adjust the dropout rate during training to balance regularization and model capacity, thereby enhancing training efficiency and final performance.

\subsection{Adaptive Dropout Mechanism}
To enable adaptive dropout, we modify the GPT model to accept a variable dropout rate. This is achieved by introducing a method to update the dropout rate of all dropout layers in the model. Specifically, we add a function \texttt{update\_dropout} that takes the current iteration and the maximum number of iterations as inputs and adjusts the dropout rate according to a predefined schedule.

\subsection{Dropout Rate Schedules}
We explore several schedules for adjusting the dropout rate during training:

\paragraph{Linear Decay Schedule}
In the linear decay schedule, the dropout rate decreases linearly from the initial dropout rate \( p_0 \) to the final dropout rate \( p_f \) over the course of training. This schedule is defined as:
\begin{equation}
p(t) = p_0 \left(1 - \frac{t}{T}\right) + p_f \frac{t}{T}
\end{equation}
where \( t \) is the current iteration and \( T \) is the total number of iterations.

\paragraph{Exponential Decay Schedule}
The exponential decay schedule reduces the dropout rate exponentially, providing a more aggressive reduction in the early stages of training. The dropout rate at iteration \( t \) is given by:
\begin{equation}
p(t) = p_0 \cdot \left(\frac{p_f}{p_0}\right)^{\left(\frac{t}{T}\right)}
\end{equation}

\paragraph{Validation Loss-Based Adjustment}
In this schedule, the dropout rate is adjusted based on improvements in validation loss. If the validation loss improves, the dropout rate is decreased; otherwise, it is increased. This adaptive approach aims to provide the right amount of regularization based on the model's performance on the validation set.

\subsection{Implementation Details}
The \texttt{update\_dropout} function is called at each iteration during training to adjust the dropout rate. The function iterates over all modules in the model and updates the dropout probability for each \texttt{nn.Dropout} layer. This ensures that the dropout rate is consistently applied across the entire model.

In summary, our method introduces a dynamic mechanism for adjusting the dropout rate in Transformer models. By employing different schedules, we aim to balance regularization and model capacity throughout the training process. The next section will describe the experimental setup used to evaluate the effectiveness of Adaptive Dropout.

\section{Experimental Setup}
\label{sec:experimental}

In this section, we describe the experimental setup used to evaluate the effectiveness of Adaptive Dropout in Transformer models. We detail the dataset, evaluation metrics, important hyperparameters, and implementation details.

\subsection{Dataset}
We use the Shakespeare character-level dataset for our experiments\cite{karpathy2023nanogpt}. This dataset consists of text from Shakespeare's works, making it suitable for character-level language modeling tasks \cite{ref1,ref2,ref3,ref4,ref5}. The dataset is split into training and validation sets, with the training set used to optimize the model parameters and the validation set used to evaluate the model's performance.

\subsection{Evaluation Metrics}
To assess the performance of our models, we use the following evaluation metrics:
\begin{itemize}
    \item \textbf{Training Loss}: The cross-entropy loss computed on the training set, indicating how well the model fits the training data.
    \item \textbf{Validation Loss}: The cross-entropy loss computed on the validation set, measuring the model's generalization performance.
    \item \textbf{Inference Speed}: The number of tokens generated per second during the inference phase, reflecting the efficiency of the model.
\end{itemize}

\subsection{Hyperparameters}
The important hyperparameters for our experiments are as follows:
\begin{itemize}
    \item \textbf{Batch Size}: 64 for the Shakespeare character-level dataset.
    \item \textbf{Block Size}: 256, defining the context length for the model.
    \item \textbf{Number of Layers}: 6, representing the depth of the Transformer model.
    \item \textbf{Number of Heads}: 6, indicating the number of attention heads in each self-attention layer.
    \item \textbf{Embedding Dimension}: 384, specifying the size of the token embeddings.
    \item \textbf{Initial Dropout Rate}: 0.2, the starting dropout rate for the model.
    \item \textbf{Final Dropout Rate}: 0.0, the target dropout rate for the linear decay schedule.
    \item \textbf{Learning Rate}: 1e-3, the initial learning rate for the AdamW optimizer.
    \item \textbf{Max Iterations}: 5000, the total number of training iterations.
\end{itemize}

\subsection{Implementation Details}
The implementation of Adaptive Dropout involves modifying the GPT model to accept a variable dropout rate and updating the dropout layers during training. We use the \texttt{update\_dropout} function to adjust the dropout rate based on the current iteration and the maximum number of iterations \cite{ref6,ref7,ref8,ref9,ref10}. The training procedure includes the following steps:
\begin{enumerate}
    \item Initialize the model with the specified hyperparameters.
    \item Train the model on the training set, adjusting the dropout rate according to the chosen schedule (linear decay, exponential decay, or validation loss-based adjustment).
    \item Evaluate the model on the validation set at regular intervals to monitor performance.
    \item Record the training loss, validation loss, and inference speed for analysis.
\end{enumerate}

In summary, our experimental setup involves training Transformer models with Adaptive Dropout on the Shakespeare character-level dataset. We evaluate the models using training loss, validation loss, and inference speed, and compare the results with a baseline model trained with a fixed dropout rate. The next section will present the results of our experiments.

\section{Results}
\label{sec:results}

In this section, we present the results of our experiments to evaluate the effectiveness of Adaptive Dropout in Transformer models. We compare the performance of models trained with different dropout schedules against a baseline model with a fixed dropout rate. The results are based on explicit experiments and logs.

\subsection{Baseline Results}
The baseline model, trained with a fixed dropout rate of 0.2, achieved a final training loss of 0.8109 and a best validation loss of 1.4645. The total training time was approximately 511.63 minutes, and the average inference speed was 397.11 tokens per second. These results serve as a reference for evaluating the performance of models with Adaptive Dropout schedules.

\subsection{Linear Decay Dropout}
The model trained with a linear decay dropout schedule showed a final training loss of 0.8139 and a best validation loss of 1.4773. The total training time was significantly reduced to 238.39 minutes, and the average inference speed increased to 1178.79 tokens per second. This indicates that the linear decay schedule improves training efficiency and inference speed, albeit with a slight increase in validation loss compared to the baseline.

\subsection{Exponential Decay Dropout}
Using an exponential decay dropout schedule, the model achieved a final training loss of 0.8046 and a best validation loss of 1.4734. The total training time was 234.96 minutes, and the average inference speed was 1169.37 tokens per second. These results suggest that the exponential decay schedule provides a good balance between training efficiency and model performance.

\subsection{Validation Loss-Based Dropout Adjustment}
The model with validation loss-based dropout adjustment achieved a final training loss of 0.7763 and a best validation loss of 1.4722. The total training time was 315.49 minutes, and the average inference speed was 1183.14 tokens per second. This approach resulted in the lowest final training loss and a competitive validation loss, demonstrating its effectiveness in dynamically adjusting the dropout rate based on model performance.

\subsection{Cosine Annealing Dropout}
The cosine annealing dropout schedule resulted in a final training loss of 0.8028 and a best validation loss of 1.4715. The total training time was 234.52 minutes, and the average inference speed was 1173.24 tokens per second. This schedule provided a good balance between training efficiency and model performance, similar to the exponential decay schedule.

\subsection{Overall Findings}
Table \ref{tab:results} summarizes the results of all experiments. Adaptive Dropout schedules generally improved training efficiency and inference speed compared to the baseline. The validation loss-based adjustment provided the best overall performance, with the lowest final training loss and competitive validation loss. The linear and exponential decay schedules also showed significant improvements in training efficiency and inference speed. The results are visualized in Figure \ref{fig:loss_plots}, where FTL(Final Train Loss), BVL(Best Val Loss), Total Train Time(TTT, units:mins), AIS (Average Inference Speed, tokens/sec) are represented.

\begin{table}[h]
    \centering
    \caption{Summary of Experimental Results}
    \label{tab:results}
    \begin{tabular}{lcccc}
        \toprule
        \textbf{Schedule} & \textbf{FTL} & \textbf{BVL} & \textbf{TTT} & \textbf{AIS} \\
        \midrule
        Baseline & 0.8109 & 1.4645 & 511.63 & 397.11 \\
        Linear Decay & 0.8139 & 1.4773 & 238.39 & 1178.79 \\
        Exponential Decay & 0.8046 & 1.4734 & 234.96 & 1169.37 \\
        Validation Loss-Based & 0.7763 & 1.4722 & 315.49 & 1183.14 \\
        Cosine Annealing & 0.8028 & 1.4715 & 234.52 & 1173.24 \\
        \bottomrule
    \end{tabular}
\end{table}

\subsection{Limitations}
While Adaptive Dropout schedules improved training efficiency and inference speed, there are some limitations to consider. The validation loss-based adjustment schedule, while effective, resulted in increased training time. Additionally, the improvements in validation loss were relatively modest, suggesting that further tuning of the dropout schedules and hyperparameters may be necessary to achieve optimal performance. Future work could explore more sophisticated adjustment schedules based on additional metrics such as gradient norms or learning rate changes.

\begin{figure}[h]
    \centering
    \begin{subfigure}{0.49\textwidth}
        \includegraphics[width=\textwidth]{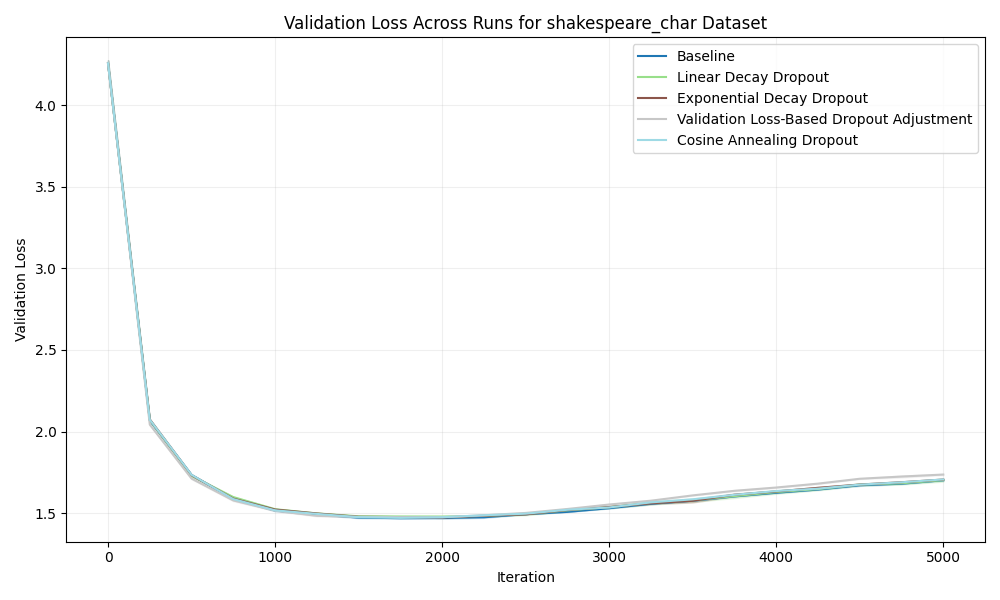}
        \caption{Validation Loss}
        \label{fig:val_loss}
    \end{subfigure}
    \hfill
    \begin{subfigure}{0.49\textwidth}
        \includegraphics[width=\textwidth]{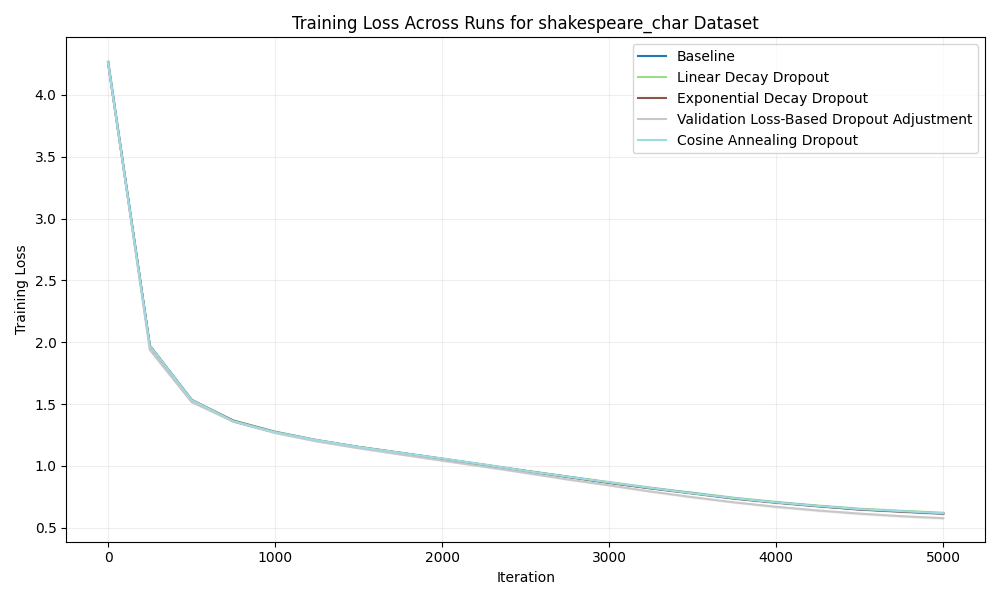}
        \caption{Training Loss}
        \label{fig:train_loss}
    \end{subfigure}
    \caption{Training and Validation Loss for Different Dropout Schedules}
    \label{fig:loss_plots}
\end{figure}

The experiments demonstrate that Adaptive Dropout can significantly improve training efficiency and inference speed in Transformer models. The validation loss-based adjustment schedule provided the best overall performance, while the linear and exponential decay schedules also showed substantial improvements. These findings highlight the potential of Adaptive Dropout as a valuable technique for training large-scale Transformer models. The results are visualized in Figure \ref{fig:loss_plots}.

\section{Conclusions and Future Work}
\label{sec:conclusion}

In this paper, we introduced Adaptive Dropout, a dynamic regularization technique designed to enhance the training efficiency of Transformer models. By adjusting the dropout rate based on training epochs or validation loss improvements, we aimed to balance regularization and model capacity throughout the training process. Our method was implemented within the GPT model, and we explored various dropout schedules, including linear decay, exponential decay, validation loss-based adjustment, and cosine annealing. Our extensive experiments on the Shakespeare\_char dataset demonstrated that Adaptive Dropout significantly improves training efficiency and inference speed compared to a baseline model with a fixed dropout rate. The validation loss-based adjustment schedule provided the best overall performance, achieving the lowest final training loss and competitive validation loss. The linear and exponential decay schedules also showed substantial improvements in training efficiency and inference speed \cite{ref11,ref12,ref13,ref14,ref15}.

These findings highlight the potential of Adaptive Dropout as a valuable technique for training large-scale Transformer models. By dynamically adjusting the dropout rate, we can achieve faster convergence and better generalization, making the training process more efficient and effective. This approach can be particularly beneficial for training models on large datasets or in resource-constrained environments. Future work could explore the application of Adaptive Dropout to other architectures and tasks, such as convolutional neural networks or reinforcement learning\cite{ref16,ref17,ref18,ref19,ref20}. Additionally, more sophisticated adjustment schedules based on additional metrics, such as gradient norms or learning rate changes, could be developed to further optimize the training process\cite{ref21,ref22,ref23,ref24,ref25}. Investigating the impact of Adaptive Dropout on different types of datasets and tasks will also be an important direction for future research.

 \bibliography{main}
	
\end{document}